\documentclass{article}
\usepackage{ijcai18}

\usepackage{times}
\usepackage{soul}
\usepackage{url}
\usepackage[hidelinks]{hyperref}
\usepackage[utf8]{inputenc}
\usepackage[small]{caption}

\usepackage{graphicx}

\title{Knowledge Amalgam: Generating Jokes and Quotes Together\thanks{\em This work was presented at 1st Workshop on Humanizing AI (HAI) at IJCAI'18 in Stockholm, Sweden.}}

\author{
Bhargav Chippada,
\space Shubajit Saha
\\ 
Microsoft R\&D\\
bhargav.chippada@microsoft.com,
shubajit.saha@microsoft.com
}

\begin{document}

\maketitle

\begin{abstract}
Generating humor and quotes are very challenging problems in the field of computational linguistics and are often tackled separately. In this paper, we present a controlled Long Short-Term Memory (LSTM) architecture which is trained with categorical data like jokes and quotes together by passing category as an input along with the sequence of words. The idea is that a single neural net will learn the structure of both jokes and quotes to generate them on demand according to input category. Importantly, we believe the neural net has more knowledge as it's trained on different datasets and hence will enable it to generate more creative jokes or quotes from the mixture of information. May the network generate a funny inspirational joke!
\end{abstract}

\section{Introduction}
Generating texts containing human sentiments is one of the most challenging tasks in natural language processing. Among the human emotions, research on the characteristics of humor, the linguistic properties of humor and its interpretation by human cognition is an active field of study.

Detailed studies have been conducted to identify the structure and principles of humor using mathematical models and generate fixed formulaic jokes using unsupervised learning from big data \cite{unjokegenbigdata}. Recurrent Neural Networks are popular for text generation tasks and they were used for humor generation \cite{neuraljoke}. The technique described in the later paper was to give the topic words (proper nouns tagged by part-of-speech tagger) as input and generate jokes on them. We believe training models with jokes and non-jokes in a supervised manner will give it more contextual data to inference from and generate creative content.

A controlled LSTM can be used to train a network in a supervised way on multiple categorical data like jokes, quotes, and tweets by augmenting the category tag to the input word at every time-step. This way the neural net can learn the difference in the semantics of a joke and quote and generate more creative content using the mix of knowledge gained by training in a supervised manner on multiple categorical data. We show how our model is able to generate a joke vs quote depending on the category input for the same prefix of words. We also found that a network trained on the combined dataset generated fewer offensive jokes compared to the one trained on just the jokes (as the jokes scraped from the internet were offensive with high probability). This is the first time anyone has used controlled LSTM architecture to generate texts with different sentiments. We show how the network learns to introduce incongruities in the generated text (making it funny) when asked to generate a joke as opposed to a quote (which is inspirational).With the current resurgence of deep-neural networks and its astounding success in natural language generation, our paper tries to achieve the above goal.

Finally, we trained our model with three categories of data namely: jokes, quotes, and tweets. We show that the network is able to generate texts belonging to the specific category when we pass the category as input along with the seed text.

\section{Related Work}
\subsection{Recurrent Neural Networks}
RNNs have been historically used for text generation. Traditional neural networks accept a fixed-sized vector as input and produce a fixed size output. But RNN is capable of accepting a sequence of vectors as input and outputs sequence of vectors. Character level RNN has been shown to generate syntactically coherent texts by \cite{sutskever2011generating} but they are not semantically meaningful since only character level information is available to the char-RNN. But encoder-decoder models \cite{sutskever2014sequence} have been proved to be successful in machine translation \cite{bahdanau2014neural} and image captioning tasks \cite{xu2015show}. Beside this vanilla-RNN suffers from the drawback that it fails to perform well in case of long-term dependencies in the text as shown by \cite{bengio1994learning}.

A variant of RNN called Long Short-Term Memory (LSTM) introduced by \cite{hochreiter1997long} have been shown to perform better than conventional RNNs by \cite{sundermeyer2012lstm} which overcomes the above-mentioned modeling limitations of RNNs. LSTMs use three gates to regulate the hidden state variable of LSTM which functions as the memory unit.

\subsection{Humor Generation}
The task of humor generation has been approached using deep neural networks with attention by \cite{neuraljoke} and unsupervised data mining to generate fixed-structure jokes by \cite{unjokegenbigdata}. In the former work, an LSTM model with attention is used to generate jokes from a dataset consisting of 7699 jokes written by a single author giving the corpus a  homogeneity of style. The jokes data is mixed with news data and a deep recurrent neural network is trained with weighted-pick strategy above the output layer to bring in randomness and a certain chance of producing funny sentences. This is the only work to the best of our knowledge which tried to mix jokes and non-jokes during training in order to bring in more information but it didn't train the network with a category tag so we have no control over what the model generates which makes judging it more subjective. It also means the network is trained in a kind of unsupervised manner when we could have trained it in a more supervised manner by telling it whether its a joke or not and later ask it to generate a joke specifically.

\section{Approach}
\subsection{Dataset}
Our training data consists of jokes, quotes, and tweets from different sources. We combined multiple sources and de-duplicated them to arrive at a large corpus for training. The two sources for jokes are CrowdTruth\footnote{\url{github.com/CrowdTruth/Short-Text-Corpus-For-Humor-Detection}} and Subreddits\footnote{\url{kaggle.com/abhinavmoudgil95/short-jokes}}. After cleaning, we ended up with 96910 jokes and a vocabulary size of 8922 words. The two sources for quotes are Quotables\footnote{\url{github.com/alvations/Quotables}} and the TheWebMiner\footnote{\url{thewebminer.com/buy-famous-quotes-database}}. After cleaning, we ended up with 43383 quotes and a vocabulary size of 8916 words. We downloaded the scraped tweets from kaggle\footnote{\url{kaggle.com/kazanova/sentiment140}} and ended up with 130250 tweets with a vocabulary size of 10805 words after cleaning. We constrained the vocabulary to about 10000 words in each case. Finally, we combined the jokes, quotes, and tweets along with their class labels (joke is 0, quote is 1, tweet is 2) into a single unified dataset. The combined dataset consists of 270543 sentences and a vocabulary size of 12614 words. Each sentence starts with a 'sos' tag and ends with a 'eos' tag to denote the start and end of sentences. The final datasets can be found on our github\footnote{\url{github.com/bhargavchippada/humor-generation/tree/master/datasets}} repository. When we train the controlled LSTM with the combined data, we use weighted sample strategy so that the three categories contribute equally to loss even though their numbers are different.

\subsection{Controlled LSTM Architecture}
\begin{figure}[h]
 \includegraphics[scale=0.51]{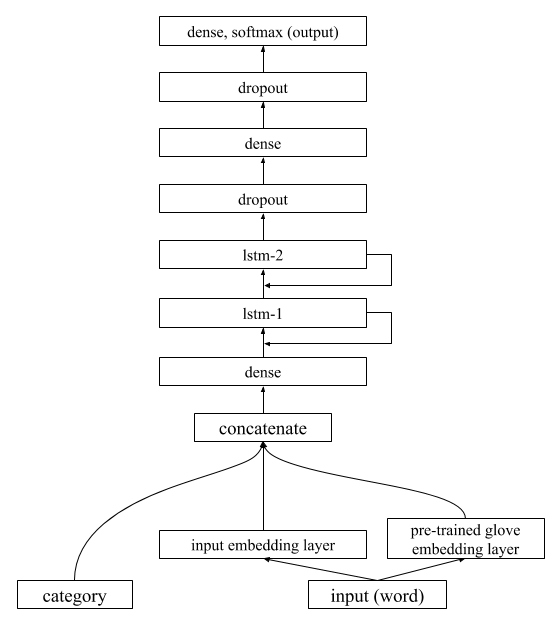}
 \caption{Controlled Stacked LSTM}
 \label{controlled_lstm}
\end{figure}

We decided to use word-level LSTM so that the network only has to learn the semantics of a sentence and not the structure of words as is the case for a char-RNN network. We tokenized each of our training examples into words and punctuation.\vphantom{As shown in figure \ref{lstm_types},We had the choice of training a many-to-one or many-to-many recurrent neural network. We choose many-to-many model so that the LSTM has more output to learn from.} At every time step of the recurrent neural network, we pass the one-hot encoded input word and the one-hot encoded sentence category. Also, we use pre-trained GloVe \footnote{\url{nlp.stanford.edu/projects/glove/}} vectors of 200 dimensions for word representation as one of the embedding layers. Figure \ref{controlled_lstm} shows the final architecture we used for training a single deep learning neural network with jokes, quotes, and tweets together with category tag and were able to successfully generate text with corresponding category later by setting the category input.


\begin{figure}[h]
 \includegraphics[scale=0.44]{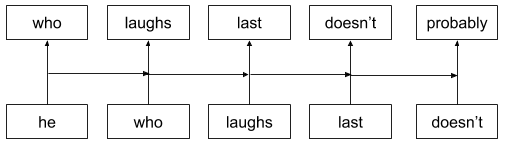}
 \caption{Sequence to Sequence}
 \label{stos}
\end{figure}

\begin{figure}[h]
 \includegraphics[scale=0.44]{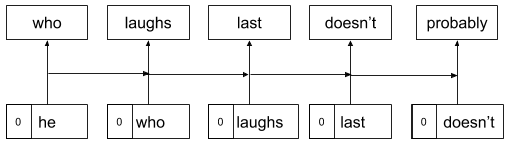}
 \caption{Sequence to Sequence with Category Tag}
 \label{stoscat}
\end{figure}
 
\subsection{Model Parameters}
Our final model has the following number of neurons at each layer. We used dropout and regularization to avoid over-fitting so we observed the network size to not have any significant effect upon increasing.
\begin{itemize}
    \item Vocabulary size: 12614
    \item GloVe embedding dim: 200
    \item Input embedding dim: 512
    \item First dense layer dim: 512
    \item LSTM-1 hidden layer dim: 1024
    \item LSTM-2 hidden layer dim: 512
    \item Final Dense layer dim: 512
    \item DropOut factor: 0.2
    \item Regularization factor: 0.00001
    \item Input sequence length: 13
\end{itemize}

\section{Experiments}
To show that controlled LSTMs work, we conducted three main experiments. The first experiment was to train the network with just jokes for a baseline model. Second, we trained the model with the words of the jokes in the correct order (tag = 0) and reverse order (tag = 1) to show that the LSTM will learn to generate jokes in forward and reverse order according to the input category tag as an extreme case. Finally, our third experiment was training the network with all three categories (jokes, quotes, and tweets) to show how the network learns to generate different emotions in the text when we input same prefix text but different category.

\subsection{Just Jokes}
The first experiment was training the model with just jokes. 
\\
\\
Examples of generated jokes: 
\begin{itemize}
\item i had to use a new word. plagiarism!
\item i had to use a lot of money in the closet. i just got laid by a chick.
\item yo mama is so hairy that bigfoot is seeing her last night.
\end{itemize}

\subsection{Forward and Reverse Jokes}
To validate our controlled model approach, we use the jokes dataset and create another reversed jokes dataset which essentially consists of every joke in reversed order of words. We train our LSTM model with the combined data with control bit set different for the two instances and later ask it to generate jokes in the required order by setting the control bit. We found that the LSTM was surprisingly good at learning the semantics of correct and reversed sentences.
\\
\\
Example reversed joke: eos . it smell can you ? horny and hungry being between difference the what's sos
\\
\\
In the correct order: sos what's the difference between being hungry and horny ? you can smell it . eos
\\
\\
Importantly, the sentence is semantically correct in the reverse order meaning the LSTM learns both the forward and reverse semantics and generates text in the required order depending on input category/control bit.
\subsection{Jokes, Quotes, and Tweets}
\begin{figure}[h]
 \includegraphics[scale=0.36]{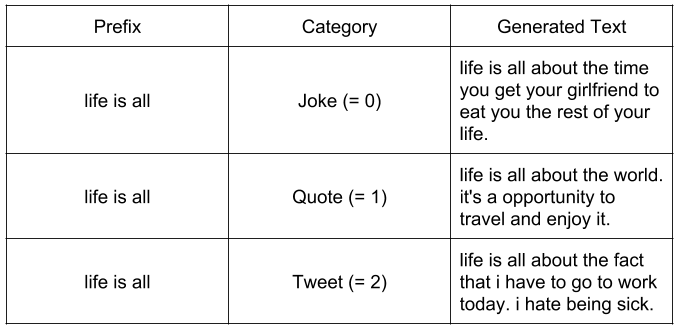}
 \caption{Generating Joke, Quotes, and Tweets}
 \label{example2}
\end{figure}

\begin{figure}[h]
 \includegraphics[scale=0.36]{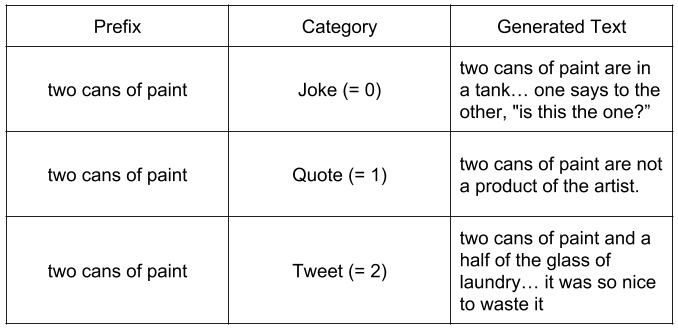}
 \caption{Generating Joke, Quotes, and Tweets}
 \label{example3}
\end{figure}

Finally, we trained the network with all three types: jokes, quotes, and tweets with their corresponding category tags (joke = 0, quote = 1, tweet = 2) as an auxiliary input. We found the network to be able to generate text with target category depending on the input tag and same seed words. Also, we found the jokes generated by this model to be far less offensive compared to the jokes generated by a network trained on just humor dataset proving the mixture of information hypothesis. Figure \ref{example1}, figure \ref{example2} and figure \ref{example3} show examples of sentences with different sentiments generated on different category inputs. While generating new texts, we use randomization to make our generated texts more diverse. We use a parameter named exploration factor to decide whether to pick the next word with the highest probability as predicted by the model or to pick a word with probabilities equal to the output of softmax layer. If exploration factor is 0 then we will always pick the word with the highest probability else if it's 0.3 then 30\% of the times we pick a word according to softmax output layer probabilities. We found that a low non-zero exploration factor gave the best results.

\begin{figure}[h]
 \includegraphics[scale=0.36]{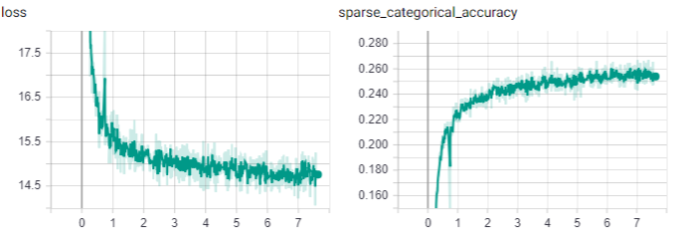}
 \caption{Loss and Accuracy for 8 epochs}
 \label{lossaccu}
\end{figure}

\section{Evaluation}
We evaluated the similarity of the generated texts with training data objectively and the humor content subjectively. We also checked the syntactic correctness of the generated sentences.

For measuring the similarity of the generated texts we used Phrase Overlap match and K-gram-Jaccard similarity as our criteria. The Phrase Overlap criterion introduced by \cite{banerjee2003extended} and extended by \cite{ponzetto2007knowledge} is chosen because it gives more weight to multi token phrase overlaps as its rarer.

\begin{equation}
    Overlap_{phrase}(s1,s2) = \sum_{i=1}^{n}\sum_{m}i^{2}
\end{equation}
\begin{equation}
    sim(s1,s2) = tanh\bigg(\frac{Overlap_{phrase}(s1,s2)}{|s1|+|s2|}\bigg)
\end{equation}

\begin{center}
\begin{table}[]
  \begin{tabular}{|c|c|c|c|}
    \hline Category & Exploration & K-Jaccard & Phrase Overlap \\ \hline
    Jokes & 0.9 & 0.09 & 0.46 \\ \hline
    Quotes & 0.9 & 0.08 & 0.43 \\ \hline
    Tweets & 0.9 & 0.09 & 0.49 \\
    \hline
  \end{tabular}
  \caption{Similarity measure with exploration = 0.9}\label{table1}
\end{table}

\begin{table}[]
  \begin{tabular}{|c|c|c|c|}
    \hline Category & Exploration & K-Jaccard & Phrase Overlap \\ \hline
    Jokes & 0.1 & 0.26 & 0.75 \\ \hline
    Quotes & 0.1 & 0.14 & 0.71 \\ \hline
    Tweets & 0.1 & 0.26 & 0.80 \\
    \hline
  \end{tabular}
  \caption{Similarity measure with exploration = 0.1}\label{table2}
\end{table}
\end{center}
The lower the $sim_{(overlap,phrase)}(s1,s2)$ value the novel the generated text which is indicative of the fact that our model is able to generalize beyond the training instances. For our experiment, we randomly sampled 100 instances from the training data and split the instances into two halves. We use the prefix half as the seed text to generate the categorical text (control tag set to 0,1 and 2). We then average the maximums of the Phrase Overlap metric between the generated texts (excluding the seed text) and all remaining training examples in our corpus to arrive at our final aggregated similarity score.

While comparing sentences with high Phrase Overlap score we found that the metric is giving a high score due to the presence of common bigram and trigram phrases. Also, the fact that there are a lot of popular phrases and templates increases the chance of a higher multigram match. For instance, there are a large number of template jokes in our corpus like "Knock Knock who's there?" or "Yo mama so fat..." so it is expected that our model will pick up these common phrases. 

For example, the generated quote "i love the music that i love because i love it . i can't tell you how to live without it" and "i am always training because i love it" has a Phrase Overlap score of 0.66 because of the presence of the four-gram phrase "because i love it" and multiple occurrences of bigram phrase "i love" but these two texts are very different semantically. Overlap of bigrams, trigrams, and even 4-grams can be expected between texts with very different meaning but phrase overlap score heavily penalizes such matches.

We also used the K-gram-Jaccard score to compute the similarity measure. We use $K=4$ to overcome the drawbacks of Phrase Overlap and compute the similarity measure of generated texts with the examples in training corpus.
\\
For two sets $A$ and $B$ Jaccard score is defined as 
\begin{equation}
    JS(A,B) = \frac{|A\cup B|}{A\cap B}
\end{equation}
We observe that with increasing exploration factor, the similarity measures decrease. This is expected because with higher exploration value we are allowing the generator to pick a not so best word which leads to more variations in the text but with lesser coherence. 

To evaluate the syntactic correctness of the generated sentences we have used the Link Grammar Parser for English language developed by \cite{grinberg1995robust}\footnote{\url{http://www.link.cs.cmu.edu/link/}} which uses the idea of linkages. A linkage is the relationship between different words in a sentence from a syntactic point of view. We use the Link Grammar Parser to find number of valid linkages of the sentence after post processing. We break the generated jokes, quotes, tweets into individual sentences. Since the parsing is dependent on capitalization and our training data has been normalized to all small letters, we add necessary capitalization before feeding sentences to the parser. Here are examples of linkage diagram produced by syntactic parsing of a generated sentence. The lines represents the links between the words and each link is labeled with the type of link. Details about the link labels can be found in the documentation\footnote{\url{http://www.link.cs.cmu.edu/link/dict/introduction.html}} of the parser.\\
\textbf{Example}: \emph{Life is all about the fact that I have to go to work today.}
\begin{figure}[h]
 \includegraphics[scale=0.25]{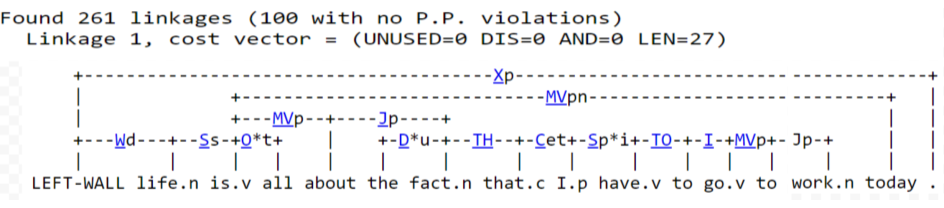}
 \caption{Parse of a generated tweet.}
 \label{example1}
\end{figure}

We briefly describe the interpretation scheme of the parser output\footnote{\url{http://www.link.cs.cmu.edu/link/explain-output.html}}. If the word in a sentence is out of vocabulary of the parser, it is followed by [?], followed by one of .n, .v, .a, or .e, depending on whether the word is being interpreted as a noun, verb, adjective, or adverb. If a word is enclosed in square brackets, this indicates that the parser was forced to delete this word  in order to find a grammatical interpretation of the sentence. The parsing occurs in multiple phases. In the first phase, attempt is made to find "complete" linkage for a sentence without leaving out any word, in which all the words are linked together. If the parser cannot interpret the sentence, it begins to relax this constraint. The value $Null count = k$ indicates that the parser is allowing $k$ words to be ignored that is, it is allowing the sentence to be partitioned into $k$ disconnected components. After finding a valid linkage, the linkage must satisfy a post-processing phase where it is evaluated against an exhaustive set of rules to be finally labeled as grammatically correct.

In the above example we can see, out of total possible 261 linkages 100 linkages were without any p.p (post processing) error. We can also see the parser identifies the parts of speech of the words and the syntactic parsing is correct.

To evaluate the overall syntactic accuracy of our corpus we consider total percentage of sentences having at least one valid linkage at Null Count 0. We generated and randomly sampled 50 quotes, jokes, and tweets each and split them into sentences. From 150 generated texts we obtained 251 sentences and processed them adequately with capitalization. We have used exploration factor 0.1 while generating the texts. The results are presented in table \ref{Syntactic-Accuracy}. The accuracy is $52\%$ i.e $52\%$ of sentences were entirely correct and $18\%$ were almost correct. Here we have to note that this numbers also includes sentences that were marked incorrect due to out of dictionary English words like \emph{iphone, dunno, gosh etc.} and proper nouns.

\begin{table}[]
\centering
\begin{tabular}{|l|l|l|}
\hline
Null Count & No. of sentences & Percentage \\ \hline
0 & 133 & 52\% \\ \hline
1 & 45 & 18\% \\ \hline
2 & 34 & 14\% \\ \hline
3 & 16 & 6\% \\ \hline
\textgreater 3 & 23 & 10\% \\ \hline
Total & 251 & 100\% \\ \hline
\end{tabular}
\caption{Syntactic Accuracy of generated sentences}
\label{Syntactic-Accuracy}
\end{table}

To evaluate the quality of the generated jokes, quotes, or tweets we rely on human judgment as there is no proven system to measure the quality of content objectively.

Giving the different category tags as input for the same prefix, we were able to generate texts of different categories.

\begin{figure}[h]
 \includegraphics[scale=0.36]{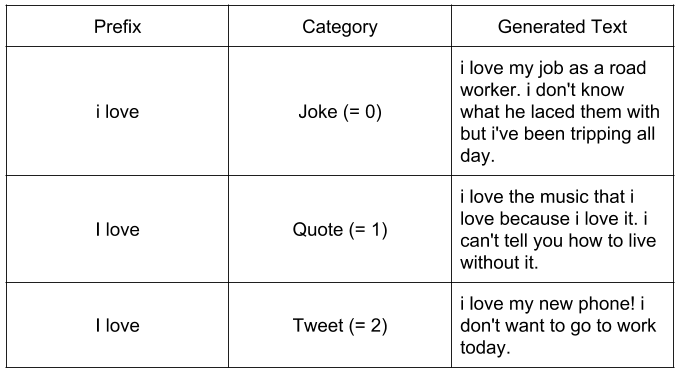}
 \caption{Generating Joke, Quotes, and Tweets}
 \label{example1}
\end{figure}

\section{Conclusion}
We have proposed a novel method of training a recurrent neural network (RNN) with categorical data to generate texts with target sentiment. We showed how our proposed controlled LSTM architecture is able to learn the semantics of different kinds of text together in a supervised manner and generate text with given sentiment on demand. The mixture of information is able to generate more creative content. We gathered a large corpus\footnote{\url{github.com/bhargavchippada/humor-generation/tree/master/datasets}} of jokes, quotes, and tweets from multiple sources for training our model which can be used for further research. The produced texts were subjectively and objectively evaluated, they are found to be semantically and syntactically coherent while expressing the required sentiment majority of the time. The neural network introduces incongruity in sentences to make them funny when asked to generate a joke, it generates inspirational quotes with meaning when asked to generate a quote, and generates casual sentences when asked to generate a tweet thus showing the network is able to learn the nature of different texts.

In future, we will build upon our idea of inferencing patterns and relations between different categories of text to generate more creative and quality content with high probability.

\bibliographystyle{named}
\bibliography{ijcai18}

\end{document}